\documentclass[10pt,twocolumn,letterpaper]{article}

\usepackage{cvpr}              

\usepackage{booktabs}
\usepackage{caption,color}
\usepackage{multirow}
\usepackage{graphicx}
\usepackage{algorithm}
\usepackage{algorithmic}
\usepackage{times}
\usepackage{pifont}
\newcommand{\cmark}{\ding{51}}
\newcommand{\xmark}{\ding{55}}
\usepackage[accsupp]{axessibility}  
\definecolor{cvprblue}{rgb}{0.21,0.49,0.74}
\usepackage[pagebackref,breaklinks,colorlinks,allcolors=cvprblue]{hyperref}

\def\paperID{66} 
\def\confName{CVPR}
\def\confYear{2026}

\title{Crowdsourcing of Real-world Image Annotation via Visual Properties}

\author{Xiaolei Diao\\
University College London\\
London, UK\\
{\tt\small xiaolei.diao@ucl.ac.uk}
\and
Fausto Giunchiglia\\
University of Trento\\
Trento, Italy\\
{\tt\small fausto.giunchiglia@unitn.it}
}

\begin{document}
\maketitle
Recent advances in data-centric artificial intelligence highlight inherent limitations in object recognition datasets.  One of the primary issues stems from the semantic gap problem, which results in complex many-to-many mappings between visual data and linguistic descriptions. This bias adversely affects performance in computer vision tasks. This paper proposes an image annotation methodology that integrates knowledge representation, natural language processing, and computer vision techniques, aiming to reduce annotator subjectivity by applying visual property constraints. We introduce an interactive crowdsourcing framework that dynamically asks questions based on a predefined object category hierarchy and annotator feedback, guiding image annotation by visual properties. Experiments demonstrate the effectiveness of this methodology, and annotator feedback is discussed to optimize the crowdsourcing setup.    

\section{Introduction}
\label{Sec:1}


In recent years, the computer vision area has experienced significant advancements, which are utilized in numerous practical application scenarios and bring convenience to daily life. As a data-driven science, machine learning models rely on high-quality datasets for training and evaluation. Driven by data-centric artificial intelligence \cite{jakubik2022data, singh2023systematic}, the importance of data quality in improving the performance of object recognition models\cite{shi2025competitive} has rekindled the interests of researchers. A critical analysis \cite{ICML-2020} of the construction process of benchmark image datasets, exemplified by ImageNet \cite{IMAGENET-2009}, reveals systemic flaws that could be a pivotal challenge in impeding the continued progress of object recognition.

The construction of non-generative image datasets in the field of computer vision typically involves crowdsourcing techniques \cite{law2011human}. During the crowdsourcing annotation process, annotators are presented with images and asked to select the most appropriate label from a set of words that are predefined categories, as introduced in some datasets, e.g., ImageNet \cite{IMAGENET-2009} and CIFAR \cite{krizhevsky2009learning}. The labelling of images in such datasets is obtained by requiring annotators to verify if the image matches a predefined category or synonym set. For instance, if the synonym set is “bear,” all images deemed by annotators to contain "bear" are categorized under this label. A widely discussed issue in this construction methodology is the many-to-many matching problem between categories and images \cite{ICML-2020, giunchiglia2023semantics}. This issue, where an image containing multiple objects is described by a single label, introduces confusing erroneous information into object recognition models, negatively impacting recognition accuracy. The advent of datasets designed for object detection tasks, such as COCO \cite{lin2014microsoft} and PASCAL VOC \cite{Everingham10}, has somewhat alleviated the problem of information clutter caused by multiple objects. The method of marking specific objects with polygonal boxes allows images to be labelled one-to-one. However, the approach to single-object category labelling remains unchanged, with the annotation process still matching images to predefined words.

\begin{figure}[t]
	\centering
	\includegraphics[width=1\linewidth]{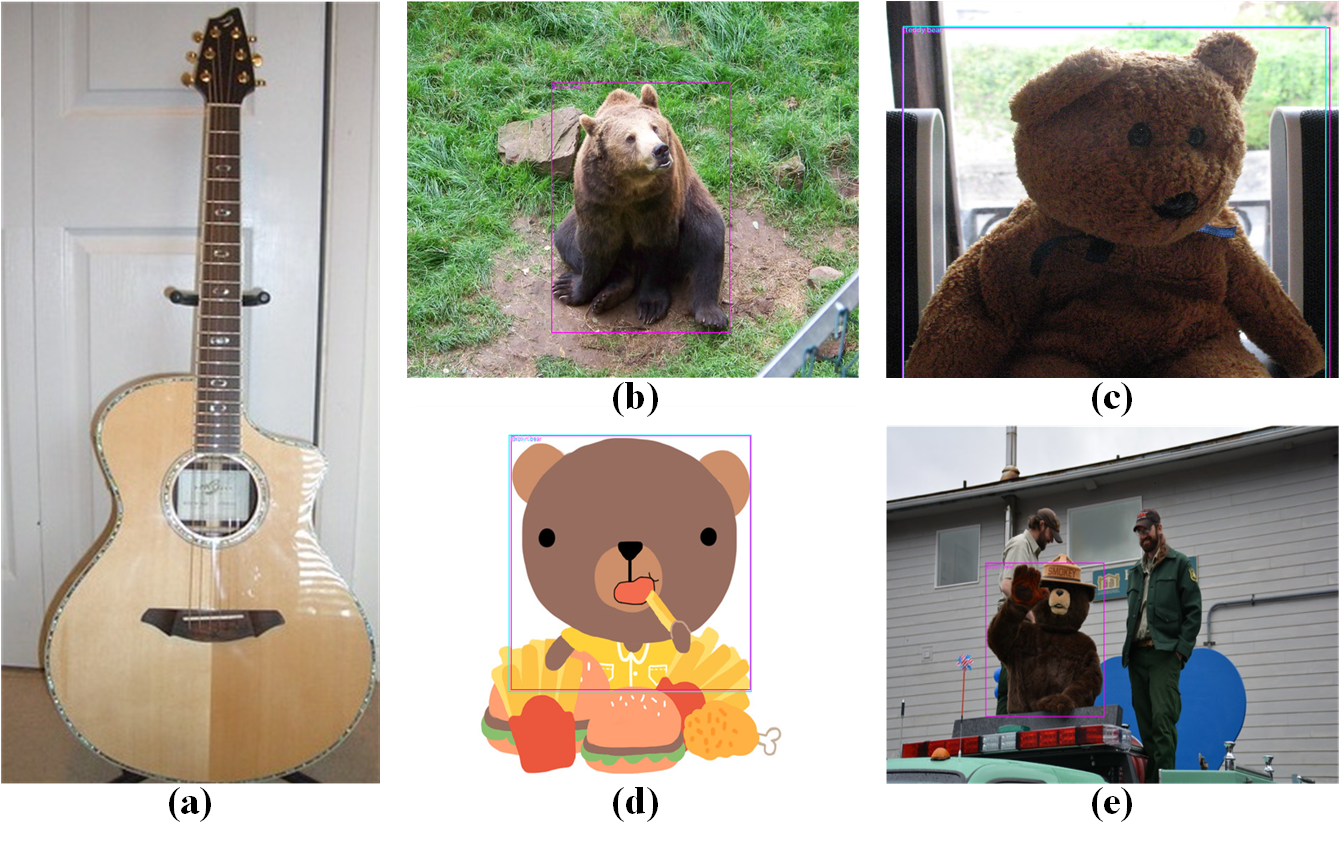}
    \caption{Images and their labels in existing image datasets. (a) This image appears in three categories of ImageNet, i.e., ``Musical Instrument", ``Guitar", and ``Acoustic Guitar". (b)-(e) These four images are all labelled as the ``Brown Bear" category in the Open Image Dataset.}
    \label{fig:1}
\end{figure}

We observe that the process of annotating these image datasets is somewhat subjective, relying on the annotator's interpretation to label the visual content. This subjectivity introduces a series of inconsistencies, as different annotators may perceive and label the same visual content with different criteria and in various ways. We give some examples in Figure~\ref{fig:1}. We found that Figure~\ref{fig:1}(a) in ImageNet falls under three categories, being labelled as ``Acoustic Guitar", ``Guitar", and ``Musical Instrument". In our opinion, this situation occurs due to the level of categorization granularity provided subjectively by the annotators. While we cannot declare that these labels are incorrect, it is undeniable that the fine-grained inconsistency in image categories may negatively impact object recognition results. In addition, in the Open Images Dataset \cite{OpenImages2}, we discovered that Figure~\ref{fig:1} (b), (c), (d) and (e) are all labelled under the category ``Brown Bear", despite significant differences: Figure~\ref{fig:1} (b) is an \textit{animal}, Figure~\ref{fig:1} (c) is a \textit{toy} and also labelled as ``Teddy Bear", Figure~\ref{fig:1} (d) is a stick figure of a \textit{cartoon} bear, and Figure~\ref{fig:1} (e) looks like a person playing the role of a brown bear doll. The reason these four images are categorized under the same label is due to the lack of constraints in upper-layer domain categories \cite{giunchiglia2023aligning}, allowing annotators to subjectively classify at their discretion. Such subjective annotation introduces significant challenges in training machine learning models, which require precise and consistent data for optimal performance.

The complexity and polysemy of natural language indeed introduce ambiguity when describing the visual information in images and their semantic interpretation, leading to the well-known issue of the semantic gap problem (SGP)\footnote{The SGP is the lack of coincidence between the information that one can extract from the visual data and the interpretation that the same data have for a user in a given situation.}\cite{SGP-2000}.
This gap exacerbates the challenge of annotation inconsistency in image datasets. A fundamental issue with existing annotation methodologies is the lack of a clearly defined process for categorizing and standardizing image annotations. As a result, there is a pressing need for an image annotation methodology that minimizes the impact of subjective interpretations by annotators, aims to ensure consistency across annotators with diverse backgrounds, and maintains high quality in image datasets.

To address these issues, we propose a new methodology for dataset construction that combines the strengths of knowledge representation, natural language processing, and computer vision technologies within a crowdsourcing framework, aiming to significantly improve the quality of image datasets. The integration of multiple technologies is intended to create a more systematic and objective method of image annotation. Specifically, we utilize knowledge representation technology\cite{shi2020learning} to construct a hierarchy of object categories\cite{diao2025semantics}, which helps establish and define the relationships between different visual elements and their semantic labels. For each category within the hierarchy, we define visual features that can determine its parent categories and distinguish it from sibling categories, presented through structured natural language. This will assist annotators in better understanding and standardizing the annotation process.

During the construction of the dataset, we rely on crowdsourcing technology, a powerful tool for collecting large-scale data, to collect labels for the image dataset. To fully leverage its potential, guiding the crowdsourcing process through explicit protocols and quality control measures is vital. In this paper, we explore how to effectively implement a defined integrated annotation method in a crowdsourcing environment, thereby providing more reliable and consistent datasets for machine learning applications in object recognition. Additionally, our dataset will eventually acquire multi-level labels, including fine-grained category labels at different levels, visual property labels, and natural language descriptions of visual features. This will make the dataset not only suitable for object recognition but also applicable to multiple tasks in the vision and language fields, including fine-grained image recognition, zero-shot image recognition, image captioning, image generation, etc.

The remainder of the paper is organized as follows. Section 2 introduces the overall methodology we proposed for image dataset construction. We illustrated the detailed labeling strategy based on crowdsourcing in Section 3. Detailed experiments and analysis of the dataset quality evaluation are given in Section 4. Section 5 gives the related work, and Section 6 concludes the paper.


\section{Image Annotation Methodology}
\label{Sec:2}

   


In this section, we introduce the proposed four-step image annotation strategy designed to effectively bridge the SGP in image dataset annotation processing.

\noindent \textbf{Label Definition.}
The first stage in our annotation methodology focuses on defining the label space for an image dataset. This is a critical step in ensuring dataset quality and prior work on image annotation. We start by identifying the categories of objects in the newly constructed dataset and organizing them into a hierarchy. In this process, we draw upon various knowledge bases, including WordNet \cite{PWN} and Wikipedia \cite{volkel2006semantic}, to define each category with precise phrases or sentences. Each label is precisely defined to represent a set of selected visual properties\cite{shi2024kae} that align with linguistic properties, eliminating ambiguities associated with informally defined labels. Such visual properties are selected under a set of canons \cite{kaula1980canons} that define the criteria of an object. For instance, the label “Goldfinch” is defined as “A small European finch having a crimson face and yellow-and-black wings”. This stage facilitates clear synonym identification among labels.

\noindent \textbf{Label Disambiguation.}
The next stage assigns a unique conceptual identifier to each label, resolving linguistic ambiguities in polysemous labels. 
Each label is mapped unambiguously to its visual representation using hierarchical identifiers, e.g., categories ``1-1" and ``2-5-3".
This approach facilitates the record of specific categories and their visual properties
, ensuring that each label reflects a distinct and recognizable concept in the visual data.

\noindent \textbf{Object Localization.}
The object localization step is designed to solve the many-to-many issues in image annotation. It involves identifying and labelling all objects in an image, thereby eliminating potential object ambiguities. The focus is on extracting perceivable visual attributes and gradually revealing their inherent causal factors. This step can be performed using object localization models \cite{redmon2018yolov3}. For multi-object images, performing object localization and cropping before annotation is an operation to ensure dataset quality. 

\noindent \textbf{Visual Classification.}
The final step guides image annotation by identifying visual properties that characterize the objects, aiming to resolve ambiguities. We predefined two sets of visual properties: \textit{visual genus} (shared properties among different objects) and \textit{visual differentia} (distinguishing properties among different objects within the same genus). For example, the \textit{visual genus} of ``Goldfinch" is ``Finch", which is used as its parent category, while the \textit{visual differentia} is defined as ``Crimson face and yellow-and-black wings", used to distinguish it from its sibling category and perform further classification.

Compared to current image dataset annotation methods, our methodology differs in two ways. Firstly, our methodology assumes that object hierarchies are constructed based on \textit{visual properties} rather than directly using category labels. Secondly, although datasets such as ImageNet claim to have hierarchies, these are usually used only for category organization rather than introduced in the annotation process. In contrast, our method actively employs visual property-based hierarchies during annotation, requiring that image annotations align with both their linguistic descriptions and visual properties. This structured methodology addresses the challenges brought by the semantic gap, ensuring more consistent and objective annotations in image datasets.

\section{Crowdsourcing Annotation Process}
\label{Sec:3}




Building upon the image dataset annotation methodology introduced in Section \ref{Sec:2}, this section details its implementation within a crowdsourcing framework. Our method utilizes hierarchical classification \cite{silla2011survey} and detailed analysis of visual properties of objects\cite{giunchiglia2023semantics}, engaging annotators in answering a series of questions related to the visual properties, i.e., \textit{visual genera} and \textit{visual differentia}. These questions are linked to a predefined object hierarchy encompassing all inheriting superordinate categories of the objects to be annotated.

\subsection{Label Preparation.}
Before initiating the annotation phase, the first two steps of the Image Annotation Methodology are completed, i.e., \textit{Label Generation} and \textit{Label Disambiguation}. Drawing upon established knowledge bases, we construct an object hierarchy with all object categories in a dataset. To ensure clarity and eliminate linguistic ambiguities, we define each category with a set of specific visual properties based on the guidance of canons \cite{kaula1980canons, SRR-89, giunchiglia2023aligning}. This preparation lays the groundwork for the annotation process.

\subsection{Image Preparation.}
The primary goal of this phase is to continuously supply quality images for annotation. In this phase, we screen images to provide ``good images\footnote{A ``good image" is one that distinctly features a single main object, captured from an optimal viewpoint with minimal noise or distortion, ensuring clear visibility of its defining visual properties.}" for the dataset. This includes the \textit{Object Localization} step using machine learning models \cite{redmon2018yolov3} to automatically identify and crop images, focusing on offering single-object images. In the absence of significant noise, we consider these cropped images as single-object images, even if minor parts of other objects are present. This step significantly reduces object ambiguities and lays the groundwork for finer categorization in the following phases.

\begin{algorithm}[tb]
    \caption{$L  = VisClassify(I)$}
    \label{alg:algorithm}
     $H$ = Tree of ($N$);\\
    $N= <id, vD>$; \{For a category node, $id$ is the conceptual identifier, $vD$ is the visual differentia properties.\}\\    
    \textbf{Input}: Image $I$;\\
    \textbf{Output}: Image label $L$;
    \begin{algorithmic}[1] 
        \STATE $R = getRoot(H)$;
        \IF {$askDif(I, R.vD) = False$}
            \RETURN $L = ``Discharged"$;    
        \ENDIF
        \STATE $L = R.vD$;
            \WHILE{$hasChild(R) = True$}
                    \WHILE{$C \in getChild(R)$}
                        \IF {$askDif(I, C.vD) = True$}
                        \STATE $R = C$;   $break$;                \ENDIF
                    \ENDWHILE
                \STATE $L = R.vD$;   
            \ENDWHILE
            \RETURN $L$;
    \end{algorithmic}
    \label{alg:1}
\end{algorithm}

\subsection{Crowdsourcing Annotation Collection.}
Once the preparation of labels and images is complete, annotators proceed with the \textit{Visual Classification}. Given the diverse backgrounds of workers\footnote{We refer to the crowdsourcing workers as ``annotators" in the following paper.} on the crowdsourcing platform, we optimize human-machine interaction during the annotation process to obtain high-quality image annotations. The hierarchy of all categories in the dataset to be annotated is provided to the annotators, along with the \textit{visual genus} and \textit{visual differentia} of each node within the hierarchy. The hierarchy is displayed interactively, allowing easy viewing and navigation of subcategories. The interface also presents the image to be annotated alongside questions related to its visual properties. The series of questions, designed to refine the labels of images iteratively \cite{ giunchiglia2023incremental}, are presented in an automated process based on the object hierarchy and the annotator's responses. These new questions are presented through a systematic process with a series of vertical and horizontal loops, aiming to iteratively refine the labels of the images being annotated. The specific procedure is outlined in Algorithm~\ref{alg:1}, which details the step-by-step approach to this iterative refinement.

In this process, the vertical loop aims to refine the labels of input images iteratively, moving from root to leaf nodes in the object hierarchy to accurately categorize them. By asking annotators if the image shares a \textit{visual genus} with specific categories in the hierarchy, candidate categories are determined, and then triggering the corresponding horizontal loop. The goal of the horizontal loop is to ascertain the most refined category within its domain for the input image, involving comparisons with subcategories of the candidate. Human annotators are asked to identify \textit{visual differentia} to determine if the object belongs to a specific subcategory. These interrelated loops, navigating through the object hierarchy, ensure the most accurate and reliable labelling.

\subsection{Quality Control.}
During the annotation phase, we implemented a strategy where each image was annotated by three different individuals. This approach was designed to enhance the reliability of perspectives in the annotation process. In the quality assessment phase, labels that received consistent annotations from at least two out of the three annotators were adopted as the final label for the dataset. In cases where there was no consensus among the initial three annotators, a fourth annotator was introduced to reassess and provide an additional annotation for the image. This step was crucial in ensuring the accuracy and consistency of the labels in our dataset, thereby enhancing the overall quality of the annotated data.

This innovative crowdsourcing methodology for image dataset annotation integrates the efficiency of machine learning with human cognitive abilities, creating a dynamic and continually improving annotation process. Guided by the object hierarchy and emphasizing the role of visual properties, this methodology ensures higher precision and consistance in image annotation, vital for advancing machine learning and AI capabilities in object recognition.

\section{Experiments and Discussion}
\label{Sec:4}

\begin{table*}[!t]
\centering
	\resizebox{1\linewidth}{!}{%
\begin{tabular}{@{}c|ccccccccccccc|c@{}}
\toprule
Categories         & I  & II  & III & IV  & V   & VI  & VII & VIII & IX & X   & XI & XII & Unrecognised            & Alpha                  \\ \midrule
Annotators group 1 & 98 & 101 & 86  & 102 & 97  & 103 & 92  & 95   & 96 & 96  & 90 & 95  & \multicolumn{1}{c|}{49} & \multirow{3}{*}{0.934} \\
Annotators group 2 & 91 & 105 & 93  & 98  & 101 & 98  & 91  & 94   & 87 & 104 & 88 & 92  & \multicolumn{1}{c|}{58} &                        \\
Annotators group 3 & 95 & 100 & 91  & 101 & 99  & 102 & 95  & 92   & 90 & 98  & 93 & 98  & \multicolumn{1}{c|}{46} &                        \\ \bottomrule
\end{tabular}}
\caption{The number of images for categories we obtained by the proposed image annotation methodology from three groups of annotators, where "I, II, ..., XII" denote twelve categories. The ``Unrecognised" indicates instances where annotators did not annotate the image into any of the twelve specific categories but rather assigned it to an upper-level category, e.g., its parent category. "Alpha" refers to Krippendorff's alpha measure, a statistical tool for assessing the reliability of these annotations.}
\label{tab:1}
\end{table*}

\begin{table*}[!t]

\centering
	\resizebox{1\linewidth}{!}{%
\begin{tabular}{@{}c|ccc|ccc@{}}
\toprule
\multirow{2}{*}{Image annotation methodology} & \multicolumn{3}{c}{50 images per task}      & \multicolumn{3}{|c}{100 images per task}     \\ \cmidrule(l){2-7} 
                                              & Alpha & Time (min) & Payments(£/p) & Alpha & Time (min) & Payments(£/p) \\ \midrule
Method A                                      & 0.912 & 4.12       & 1             & 0.896 & 8.27       & 2             \\
Method B                                      & 0.937 & 5.08       & 1.5           & 0.914 & 9.22       & 3             \\
Method C (Ours)                               & 0.974 & 5.47       & 1.5           & 0.958 & 9.47       & 3             \\ \bottomrule
\end{tabular}}
\caption{Results of implementing three different image annotation methodologies, including inter-class agreement evaluation with Krippendorff's alpha measure, time cost on single tasks, and payment to one annotator for one task. To evaluate the proper number of images per task, we present a comparison of results for tasks with 50 images and 100 images.}
\label{tab:2}
\end{table*}

To validate the effectiveness of our proposed image annotation methodology within a crowdsourcing framework, a series of experiments were conducted focusing on machine learning results and assessing the consistency among annotators as measures of dataset quality.

\subsection{Experimental Setup}
\noindent \textbf{Image and Labels Collection.}
A total of 1200 images were collected, representing 12 different object categories. These categories spanned three domains: birds (5 categories), vehicles (3 categories), and musical instruments (4 categories). Following the guidance of WordNet and Wikipedia, these object categories were organized into a hierarchy. For each node (category) in the hierarchy, a pair of \textit{visual genus} and \textit{visual differentia} was defined to guide the annotation process.

\noindent \textbf{Crowdsourcing development.}
A new platform\footnote{The platform will publish, currently anonymous due to blind review.} was developed and linked to a crowdsourcing platform. 
Building upon the established image annotation methodology, we factored in the flexibility of task deployment as a critical aspect of our crowdsourcing approach. To optimize this, we selected Prolific\footnote{https://www.prolific.com/.} as our primary crowdsourcing platform for disseminating annotation tasks. Every response from the annotators was recorded on our platform. The selection of image labels is based on the appropriate category level within the dataset's hierarchy, aligning with the specific requirements of the dataset. Typically, the most fine-grained annotation level is selected as the label for the image. 

\subsection{Inter-annotator agreement.} 
We introduce Krippendorff's Alpha\cite{hughes2021krippendorffsalpha} for measuring inter-annotator agreement. We assign 1200 images belonging to twelve object categories to three groups of annotators for annotation. Each annotator has seen each image only once, and every image was annotated by three different annotators. Upon completion of the image annotation, the number of images in each category is as shown in Table \ref{tab:1}. In the consistency evaluation, the result as high as 0.934 indicates a significant level of agreement among the three groups of annotators. This high consistency validates the reliability of our method in obtaining image annotations.

\subsection{Experiments on crowdsourcing setting} 
Before deploying large-scale image annotation with crowdsourcing, it's crucial to establish the optimal experimental setup for crowdsourcing to ensure high-quality image annotations. To this end, we designed a series of experiments to collect image annotations under various crowdsourcing experimental parameter settings. The goal of these experiments is to identify the most effective interaction settings for the crowdsourcing framework by assessing the quality of the annotations and analyzing feedback from annotators.

\subsubsection{Experiment 1: Annotation Methodology Evaluation.}
This set of experiments is designed to evaluate the quality of the image annotation results by comparing different annotation methodologies. There are three different annotation methodologies, including:
\begin{itemize}
\item Method A: Name labels only. Annotators are provided with only the name (present as words or phrases) of the category to label the images, as the existing annotation methodology used, e.g., ImageNet.
\item Method B: Name labels in an object hierarchy. This method involved using name labels within a predefined hierarchy for image annotation.
\item Method C: Visual properties labels in an object hierarchy. Annotators are provided a set of visual properties, including \textit{visual genus} and \textit{visual differentia} for every category within a predefined hierarchy to label images, namely our proposed methodology.
\end{itemize}
The results are shown in Table~\ref{tab:2}. Method A resulted in quicker annotations but had lower inter-class agreement. This illustrates that Method A relies heavily on semantic processing, which can be subjective and ambiguous, leading to the semantic gap among annotators. 
Method B improved the inter-class agreement among annotators over Method A, and Method C demonstrated the highest accuracy and consistency among the three methods, though it required more time from the annotators. The structured approach in Methods B and C reduces cognitive load by providing a clear framework for annotation. However, Method C further reduces cognitive ambiguity by integrating visual properties, leading to higher-quality annotations. Cognitive science suggests that deeper processing of information, like analyzing \textit{visual genus} and \textit{visual differentia}, leads to better memory and understanding. This is demonstrated in the higher agreement of Method C.
This experiment demonstrates the importance of annotation methods based on descriptions of structured visual properties in improving the quality of datasets. Our proposed image annotation methodology, although more time-consuming, provides a more robust framework for accurate image annotation. This insight is crucial for designing annotation tasks, especially in domains that require high annotation accuracy and consistency.

\subsubsection{Experiment 2: Costs Efficiency Analysis.}
In this experiment, we continue to compare three image annotation methodologies, as introduced in Experiment 1, to evaluate their cost efficiency. The methodologies were assessed based on their annotation consistency (measured using Krippendorff's Alpha), the time taken to annotate a set number of images, and the payments made to annotators. The results for both 50 and 100 image tasks were analyzed and the results are shown in Table~\ref{tab:2}.
We can find that:
\begin{itemize}
\item Time vs Quality Trade-off: Our proposed methodology, namely Method C, consistently outperformed the others in terms of annotation consistency across both tasks. Although it required slightly more time and payment, the increase in quality as evidenced by the higher Alpha values suggests a valuable trade-off.
\item Learning Curve Efficiency: By observing the two sets of tasks with  50 to 100 images, we found an increase in annotation speed. We think this is because annotators became more familiar with our hierarchical structure, suggesting a decrease in time and cost as efficiency improves over time.
\item Cost Implications: The higher consistency achieved with our methodology suggests that the increased cost per image is a worthwhile investment, especially for applications where annotation accuracy is critical.
\end{itemize}
As a result, while our annotation methodology requires a bit more time and incurs a slightly higher cost than the existing annotation methodology, the superior consistency and accuracy of annotations justify these additional resources. This consideration is particularly crucial in contexts where the quality of data is paramount.

\subsubsection{Experiment 3: Number of Images per Task.}
This experiment is designed to explore the balance between task length and annotator motivation in crowdsourcing tasks. We varied the number of images per task, including 10, 30, 50, and 100, and collected feedback from the annotators on their experience.
Feedback from annotators indicated that tasks with too few images (like 10) were quickly completed but resulted in low compensation,  reducing their appeal. Conversely, tasks with a high number of images led to boredom and decreased enthusiasm. Most annotators (8 in 10) agree that a task comprising 50 images is best.

The above feedback is in line with our general understanding. Tasks with too many images can lead to cognitive overload. Attention tends to wane over time with repetitive tasks, impacting the quality of annotations. In contrast, tasks with too few images may not engage the annotators sufficiently, leading to a lack of focus or investment in the task. The balance among task size, time investment, and compensation plays a crucial role in maintaining annotator motivation. A moderate number of images (like 50) in a task keeps annotators engaged without overwhelming them, facilitating sustained attention and higher-quality annotations.

\subsection{Downstream Visual Recognition Evaluation.}

Beyond annotation consistency and cost-efficiency, we further examine whether the proposed annotation design provides more effective supervision for downstream visual recognition. To this end, we report complementary results from a previously completed controlled evaluation on an ImageNet-derived benchmark, namely name-based labels (Method A) and a dataset constructed with the same methodological principles as those adopted in this paper (Method C). The two datasets span the same three domains considered in this work, namely birds, vehicles, and musical instruments.

All models were trained and evaluated under identical settings so that the effect of annotation design could be isolated from other factors. Following the protocol in the prior study, 80\% of the images were used for training and the remaining 20\% for testing. The experiments were implemented in PyTorch with the same data augmentation strategy, including horizontal flipping, random scaling, and random $224 \times 224$ cropping. Optimization was performed with Adam, using an initial learning rate of 0.0002, momentum of 0.9, and weight decay of $10^{-8}$. All evaluated models were initialized from ImageNet pre-training.

The results are reported in Table~\ref{tab:downstream_results}. Across all eight evaluated backbones, the refined annotation protocol consistently outperforms the conventional name-label baseline. Specifically, AlexNet improves from 0.543 to 0.596, ZFNet from 0.612 to 0.657, VGG from 0.655 to 0.743, and GoogleNet from 0.734 to 0.835. 

These results provide downstream evidence that the advantages of the proposed methodology extend beyond higher inter-annotator agreement. In the crowdsourcing experiments reported earlier, Method C achieved the highest Krippendorff’s Alpha under both the 50-image and 100-image task settings, demonstrating that hierarchy-guided annotation with explicit visual properties produces more consistent labels. The downstream evaluation shows that such improvements in annotation quality also translate into stronger supervision signals for visual recognition models. Taken together, these findings support the same conclusion: annotation protocols that make category assignment more explicitly grounded in structured visual semantics yield labels that are not only more reliable for human annotators but also more beneficial for subsequent computer vision tasks.

\begin{table}[t]
\centering
\setlength{\tabcolsep}{4.5pt}
\resizebox{1\linewidth}{!}{
\begin{tabular}{lccc}
\toprule
\textbf{Model} & \textbf{Method A} & \textbf{Method C (Ours)} & \textbf{Improvement} \\
\midrule
AlexNet \cite{alexnet}           & 0.543 & 0.596 & 9.76\%  \\
ZFNet \cite{zfnet}               & 0.612 & 0.657 & 7.35\%  \\
VGG \cite{vgg}                   & 0.655 & 0.743 & 13.44\% \\
GoogleNet \cite{googlenet}       & 0.734 & 0.835 & 13.76\% \\
ResNet \cite{resnet}             & 0.593 & 0.732 & 23.44\% \\
DenseNet \cite{densenet}         & 0.724 & 0.793 & 9.53\%  \\
RAN \cite{ran}                   & 0.713 & 0.784 & 9.96\%  \\
SENets \cite{senet}              & 0.728 & 0.811 & 11.40\% \\
\bottomrule
\end{tabular}}
\caption{Complementary downstream visual recognition results on an ImageNet-derived benchmark annotated under two different designs: a conventional name-based labeling condition and the full visual-property-guided design. Higher values indicate better classification accuracy.}
\label{tab:downstream_results}
\end{table}

\begin{table*}[t]
\centering
\setlength{\tabcolsep}{5pt}
\resizebox{1\linewidth}{!}{
\begin{tabular}{lcc|ccc|ccc}
\toprule
\multirow{2}{*}{\textbf{Method}} & \multirow{2}{*}{\textbf{Hierarchy}} & \multirow{2}{*}{\textbf{Visual Properties}} & \multicolumn{3}{c|}{\textbf{50 Images per Task}} & \multicolumn{3}{c}{\textbf{100 Images per Task}} \\
\cmidrule(lr){4-6} \cmidrule(lr){7-9}
 &  &  & \textbf{Alpha} & \textbf{Time (min)} & \textbf{Payment (£/p)} & \textbf{Alpha} & \textbf{Time (min)} & \textbf{Payment (£/p)} \\
\midrule
Method A & \xmark & \xmark & 0.912 & 4.12 & 1.0 & 0.896 & 8.27 & 2.0 \\
Method B & \cmark & \xmark & 0.937 & 5.08 & 1.5 & 0.914 & 9.22 & 3.0 \\
Method C (Ours) & \cmark & \cmark & 0.974 & 5.47 & 1.5 & 0.958 & 9.47 & 3.0 \\
\midrule
$\Delta$ (B $-$ A) & -- & -- & +0.025 & +0.96 & +0.5 & +0.018 & +0.95 & +1.0 \\
$\Delta$ (C $-$ B) & -- & -- & +0.037 & +0.39 & +0.0 & +0.044 & +0.25 & +0.0 \\
$\Delta$ (C $-$ A) & -- & -- & +0.062 & +1.35 & +0.5 & +0.062 & +1.20 & +1.0 \\
\bottomrule
\end{tabular}}
\caption{Ablation study on annotation design under the crowdsourcing setting. Method A uses category names only; Method B adds an object hierarchy; Method C further adds visual properties (visual genus and visual differentia) within the hierarchy. Higher Alpha indicates better inter-annotator agreement.}\label{tab:ablation_annotation}
\end{table*}

\subsection{Ablation Study on Annotation Design.}

To better understand which design choices are responsible for the quality improvement of the proposed annotation methodology, we reformulate the comparison among Method A, Method B, and Method C as an ablation study on annotation design. The three methods differ only in how semantic guidance is provided to annotators. Method A uses category names only, Method B introduces category names within a predefined object hierarchy, and Method C (our method) further incorporates explicit visual properties, i.e., visual genus and visual differentia, within the same hierarchy. In this sense, the transition from Method A to Method B isolates the effect of hierarchical organization, while the transition from Method B to Method C isolates the additional effect of grounding the annotation process in structured visual properties. The image set, crowdsourcing platform, and evaluation criteria remain the same across the three methods, allowing the observed differences to be attributed directly to the annotation design.

The results are reported in Table~\ref{tab:ablation_annotation}. A clear and consistent trend can be observed across both task settings. First, adding a hierarchy already improves annotation reliability. Compared with Method A, Method B increases Krippendorff's Alpha from 0.912 to 0.937 on tasks with 50 images, and from 0.896 to 0.914 on tasks with 100 images. This indicates that even when annotators are still guided by category names, organizing these labels within a structured hierarchy reduces ambiguity and provides a clearer decision path during annotation.

Second, introducing visual properties on top of the hierarchy leads to a further and larger improvement. Method C achieves the highest inter-annotator agreement in both settings, reaching 0.974 for 50-image tasks and 0.958 for 100-image tasks. Relative to Method B, the gain brought by visual-property guidance is +0.037 and +0.044 in Alpha for the two task settings, respectively. This shows that the main benefit of the proposed methodology does not come only from hierarchical organization, but from requiring annotators to justify category assignment through explicit visual evidence. In other words, visual genus and differentia provide a stronger constraint on the interpretation of the image, thereby reducing the semantic gap among annotators.

The ablation also reveals the trade-off between annotation quality and annotation effort. Method C requires slightly more time than the other two methods, but the extra overhead is modest. For example, compared with Method B, Method C requires only 0.39 additional minutes for 50-image tasks and 0.25 additional minutes for 100-image tasks, while yielding the highest agreement. Moreover, for 50-image tasks, Method C achieves a substantial improvement in Alpha over Method A with only a small increase in payment (£1.5 versus £1.0) and with the same payment as Method B. A similar pattern is observed for 100-image tasks. These results suggest that the quality gain achieved by adding structured visual properties is not only statistically meaningful but also practically cost-effective.

Overall, this ablation study supports two conclusions. First, hierarchical structure is already beneficial for reducing ambiguity in image annotation. Second, the additional use of visual properties is the dominant factor that further improves annotation consistency. Therefore, the proposed Method C can be understood as the full design, where hierarchy provides the structural scaffold for decision making and visual properties provide the fine-grained semantic constraints needed for accurate and reliable annotation.

\subsection{Qualitative Analysis.}

We further provide a qualitative analysis to illustrate the type of ambiguity addressed by the proposed methodology. Existing image datasets often contain cases where one image is associated with multiple labels at different granularity levels, or where visually distinct instances are grouped under the same category. A representative example is an image that appears under the labels ``Musical Instrument'', ``Guitar'', and ``Acoustic Guitar'' in ImageNet. Another example is the ``Brown Bear'' category in existing benchmarks, where images of a real bear, a teddy bear, a cartoon bear, and a person wearing a bear costume may all be assigned the same label.

Under Method A, such images are annotated primarily through category names, which leaves considerable room for subjective interpretation. Method B reduces part of this ambiguity by organizing labels in a hierarchy, but category assignment is still driven mainly by the names of categories. In contrast, Method C requires annotators to verify explicit visual properties step by step. For instance, an image is not directly annotated as ``Acoustic Guitar''; instead, the annotator verifies whether the object is a musical instrument, then a string instrument, then a guitar, and finally whether it satisfies the visual differentia corresponding to ``Acoustic Guitar''. This process reduces ambiguity caused by inconsistent granularity, incomplete domain knowledge, and category-name polysemy.

Qualitative examples also help explain the expected downstream gains. When the final labels are aligned more closely with the visual evidence present in the image, the resulting supervision signal becomes more discriminative. As a consequence, models trained on Method C annotations are expected to rely more on object-specific visual cues and less on incidental context. This qualitative evidence complements the quantitative results by showing how the proposed methodology addresses semantic ambiguity at the level of individual examples.

\section{Related work}
\label{Sec:5}




The problem of dataset quality has been of concern to researchers for a while. An early paper focusing on this problem is \cite{torralba2011unbiased}, which highlighted the biases inherent in popular datasets. The seminal work of \cite{SGP-2000} identified the Semantic Gap Problem (SGP) as a primary cause of quality issues in datasets, emphasizing the discrepancy between the computational understanding of images and their human interpretation.
Recent efforts have shifted towards addressing methods to tackle dataset quality problems. \cite{2021-MLDatasetDev} advocates for a more cautious approach in dataset development, paying attention to its limitations and impacts. This includes considering the diversity and representativeness of the dataset to ensure its broad applicability \cite{diao2023toward}. 
In the pursuit of fairer datasets, \cite{2020-ACMFAT} proposes a vertical method to balance the personnel subtree in ImageNet. This approach seeks to address biases and improve the representativeness of datasets. 
\cite{yun2021re} demonstrated how the quality of ImageNet could be improved by using classifiers trained on higher-quality datasets. This method leverages the strengths of superior datasets to enhance the annotations and classifications in existing datasets.
Another interesting perspective is provided by \cite{diaobuilding}, discussing the feasibility of using visual features to annotate multimedia datasets. This approach underscores the potential of incorporating more complex and nuanced visual cues in dataset annotation.
Our work complements these studies by addressing the gap left in how interactions with humans during the annotation collection process can lead to higher-quality datasets. We focus on optimizing human-machine interactions to improve the quality of annotations, considering factors like annotation methodology, task design, and annotator training. This approach is crucial for developing datasets that accurately reflect human perceptions and interpretations, thereby enhancing the reliability and utility of machine learning models trained on these datasets \cite{li2023fcc, shi2025learn}.

The quality of the annotation process in crowdsourcing communities has been a subject of widespread interest, as seen in works \cite{nowak2010reliable} and \cite{ewerth2017machines}. These studies have delved into various aspects of crowdsourced annotations, highlighting common challenges and proposing solutions to enhance data quality.
Further, \cite{daniel2018quality} offers a comprehensive description of crowdsourcing quality and extensively analyzes existing techniques. This work is instrumental in understanding the key elements that contribute to the reliability and validity of crowdsourced data.
Recently, there has been a shift towards operational improvements in the process to enhance quality. For instance, works by \cite{kyriakou2021crowdsourcing} and \cite{demartini2021managing} focus on process optimization to achieve higher quality outcomes in crowdsourced tasks.
Most pertinent to this paper, \cite{nassar2019assessing} developed and utilized a set of metrics, including Krippendorff's alpha, aimed at monitoring the image annotation process. Their approach offers valuable insights into the assessment and maintenance of annotation quality in crowdsourced projects.
The above works focus on measuring effects and controlling annotator behavior, while our emphasis differs in aligning the semantics encoded in images and NLP descriptions. Specifically, we address the challenge of the semantic gap problem and the many-to-many mappings it entails. Our work concentrates on aligning the visual content of images with their semantic interpretation in NLP, aiming to bridge the gap between human interpretation. This focus is crucial for enhancing the accuracy and relevance of annotations in datasets used for machine learning and AI applications, particularly in fields where precise semantic understanding is paramount.

\section{Conclusion}
\label{Sec:6}





In this study, we developed and validated an innovative image annotation methodology within a crowdsourcing framework, significantly enhancing annotation consistency and accuracy. Our approach, grounded in knowledge representation, natural language processing, and computer vision, focuses on hierarchical categorization and detailed visual property analysis, offering a novel solution to challenges in dataset quality and semantic interpretation. 
Finally, our dataset will encompass multi-level labels, and can be used in multiple downstream tasks across vision and language domains, including fine-grained and zero-shot image recognition, image captioning, and image generation. 
Future research directions include developing automated and adaptive systems, exploring cross-domain applications, and enhancing training protocols. 

\section*{Acknowledgements}
The work described in this presentation has been conducted within the project TRUMAN. The research leading to these results has received funding from HORIZON-CL4-2024-HUMAN-03, under Grant Agreement no 101214000.

{
    \small
    \bibliographystyle{ieeenat_fullname}
    \bibliography{main}
}


\end{document}